\documentclass[10pt,conference,a4paper]{IEEEtran}
\usepackage{ifpdf}

\ifCLASSINFOpdf
  \usepackage[pdftex]{graphicx}
  \usepackage{graphicx, array, blindtext}
  \graphicspath{{../pdf/}{../jpeg/}}
  \DeclareGraphicsExtensions{.pdf,.jpeg,.png}
\else
  \usepackage[dvips]{graphicx}
  \graphicspath{{../eps/}}
  \DeclareGraphicsExtensions{.eps}
  \fi
\usepackage{amsmath,amsxtra,amssymb,amsthm,latexsym,amscd,amsfonts}
\usepackage[utf8]{vntex}
\ifCLASSINFOpdf
\usepackage[pdftex]{graphicx}
\DeclareGraphicsExtensions{.pdf,.jpeg,.png,.jpg}
\else
\usepackage[dvips]{graphicx}
\DeclareGraphicsExtensions{.eps}
\fi
\usepackage{array}
\usepackage{epstopdf}
\usepackage{anysize}
\marginsize{2.4cm}{2.0cm}{3.2cm}{3.0cm}
\usepackage{multicol}
\usepackage{balance}
\usepackage{longtable}

\makeatletter
\def\ScaleIfNeeded{\ifdim\Gin@nat@width>\linewidth\linewidth\else\Gin@nat@width\fi}

\begin{document}
\columnsep=0.63cm
\def\mathbi#1{\boldsymbol{#1}}
\def\erfc{\:\mathrm{erfc}}
\def\arg{\:\mathrm{arg}}
\def\E{\:\mathrm{E}}
\def\sinc{\:\mathrm{sinc}}
\def\T{\mathrm{T}}
\def\H{\mathrm{H}}
\newcommand{\bigsize}{\fontsize{16pt}{20pt}\selectfont}

%
\include{Abbr}
\title{Phát triển nền tảng Tương tác người - Robot Tay máy đôi dựa trên ROS và Trí tuệ nhân tạo đa thể thức}

\author{
\IEEEauthorblockN{
Nguyễn Cảnh Thanh, Nguyễn Bá Phượng, Trần Hồng Quân, Đỗ Ngọc Minh, Đinh Triều Dương \\ và Hoàng Văn Xiêm
} 
\IEEEauthorblockA{ Bộ môn Kỹ thuật Robot, Khoa Điện tử - Viễn Thông, Trường Đại học Công Nghệ - Đại học Quốc gia Hà Nội\\
		Email: canhthanhlt@gmail.com, nguyenbaphuong1992@gmail.com, tranhongquan1258@gmail.com, \\dongocminh@vnu.edu.vn, duongdt@vnu.edu.vn, xiemhoang@vnu.edu.vn}
}
\maketitle

\begin{abstract}
Trong bài báo, chúng tôi đề xuất phát triển nền tảng tương tác giữa người và hệ thống Robot Tay máy đôi dựa trên Robot Operating System (ROS) và mô hình trí tuệ nhân tạo đa thể thức. Nền tảng đề xuất của chúng tôi bao gồm hai thành phần chính: hệ thống phần cứng tay máy đôi và phần mềm bao gồm các tác vụ xử lý ảnh, xử lý ngôn ngữ tự nhiên dựa trên 3D camera và máy tính nhúng. Đầu tiên, chúng tôi thiết kế và xây dựng một hệ thống Robot tay máy đôi với sai số vị trí nhỏ hơn 2cm, có thể hoạt động độc lập, thực hiện các nhiệm vụ trong công nghiệp, dịch vụ đồng thời mô phỏng, mô hình hóa robot trong hệ điều hành ROS. Thứ hai, các mô hình trí tuệ nhân tạo trong xử lý ảnh được tích hợp nhằm thực hiện các tác vụ gắp và phân loại vật thể với độ chính xác trên 90\%. Cuối cùng, chúng tôi xây dựng phần mềm điều khiển từ xa bằng giọng nói thông qua mô hình xử lý ngôn ngữ tự nhiên. Thực nghiệm chứng mình độ chính xác của mô hình trí tuệ nhân tạo đa thể thức và sư linh hoạt trong tương tác của hệ thống Robot tay máy đôi trong môi trường hoạt động với con người. 

\end{abstract}

\begin{IEEEkeywords}
Robot tay máy đôi, Trí tuệ nhân tạo, ROS, HRI.
\end{IEEEkeywords}
\IEEEpeerreviewmaketitle
\section{GIỚI THIỆU}

Hiện nay, ngành robot đang phát triển, sử dụng rộng rãi đặc biệt là robot tay máy. Việc triển khai, làm việc trong môi trường chứa nhiều rủi ro, không đảm bảo đòi hỏi sự ra đời của các loại robot có thể thực hiện các chức năng thay thế con người đồng thời có thể điều khiển thông minh từ xa. Để đối phó với các nhiệm vụ phức tạp và môi trường làm việc thay đổi, robot công nghiệp truyền thống không thể đáp ứng nhu cầu của các nhiệm vụ. So với điều này, robot tay máy đôi có không gian làm việc lớn hơn và tính linh hoạt cao hơn từ đó có thể đáp ứng tốt hơn các yêu cầu về độ phức tạp và độ chính xác cao trong các nhiệm vụ như sản xuất công nghiệp, dịch vụ, ...

Việc điều khiển robot theo cách truyền thồng đòi hỏi người điều khiển phải tốn rất nhiều thời gian có kĩ năng chuyên sâu trong kỹ thuật \cite{Makrini2017}. Tương tác người-robot phát triển thúc đẩy hiệu quả công việc dựa trên phương thức điều khiển nhanh chóng tiêu biểu là điều khiển bằng nhận dạng cử chỉ, thái độ, và mới nhất là giọng nói của con người. Các nghiên cứu chỉ ra hiệu quả vượt trội của việc sử dụng HRI thông qua màn hình, camera so với các phương pháp trước đây \cite{Radmard2015}.

Hệ điều hành ROS (Robot Operating System) là hệ điều hành mã nguồn mở phổ biến nhất trong công nghệ robot hiện nay \cite{thanh2021}. ROS cung cấp nhiều công cụ phát triển và các tệp thư viện phong phú để phát triển robot, giúp cải thiện đáng kể hiệu quả phát triển robot và tiết kiệm chi phí phát triển \cite{Cong2020}. 

Các nghiên cứu \cite{Cong2020, Xu2022, Sep2020} xây dựng hệ thống tay máy đôi trên nền tảng ROS. Nghiên cứu \cite{Xu2022} thiết kế hệ robot di động tay máy đôi từ đó tạo nên khung điều khiển thực tế đồng thời thiết lâp các ràng buộc nhằm xác minh tính khả thi của nền tảng. Nghiên cứu \cite{Cong2020} đưa ra mô hình động học của robot tay máy đôi đồng thời lập quỹ đạo dựa trên tập mô hình robot trong Moveit sau đó trực quan hóa và xác minh tính đúng đắn dựa trên Rviz.

Bài \cite{Hameed2018, minh2018, Zhang2018, Liu2021} ứng dụng trí tuệ nhân tạo trong việc điều khiển, tương tác người-robot. Nghiên cứu \cite{Hameed2018} sử dụng nhận dạng giọng nói và thiết kế giao diện nhằm điều khiển robot NAO, robot sử dụng máy ảnh để đếm số lượng mặt hướng tới nó nhằm đo sự chú ý. Nghiên cứu \cite{minh2018} đề xuất nâng cao độ chính xác trong việc xác định vị trí thực thi của robot thông qua xử lý ảnh và điều khiển robot bằng giọng nói. Nghiên cứu \cite{Zhang2018} xây dựng và điều khiển robot tay bằng giọng nói, kết quả cho thấy tính khả thi cũng như khả năng áp dụng thực tế. Nghiên cứu \cite{Liu2021} điều khiển robot hợp tác bằng học củng cố sâu từ đó hoàn thành được các nhiệm vụ hợp tác và khả năng thich ứng tốt hơn tuy nhiên yêu cầu bộ xử lý có cấu hình cao, khó có thể nhúng trực tiếp trong robot.

Trong bài báo này, chúng tôi phát triển nền tảng Robot tay máy đôi dựa trên ROS và tương tác người-robot với ba đóng góp chính. Thứ nhất, chúng tôi xây dựng được hệ thống tay máy đôi hoàn hiện bao gồm cả phần cứng và phần mềm. Tiếp theo, chúng tôi kết hợp xử lý ảnh giúp robot phân loại vật thể. Cuối cùng, hệ thống phần mềm điều khiển từ xa qua giọng nói được triển khai. Tất cả các thành phần đều được khai thác trên hệ điều hành ROS. 
%


\section{ĐỀ XUẤT HỆ THỐNG}
\label{Sec:System}
\begin{figure}[!ht]
    \centering
    \includegraphics[width=0.38\textwidth]{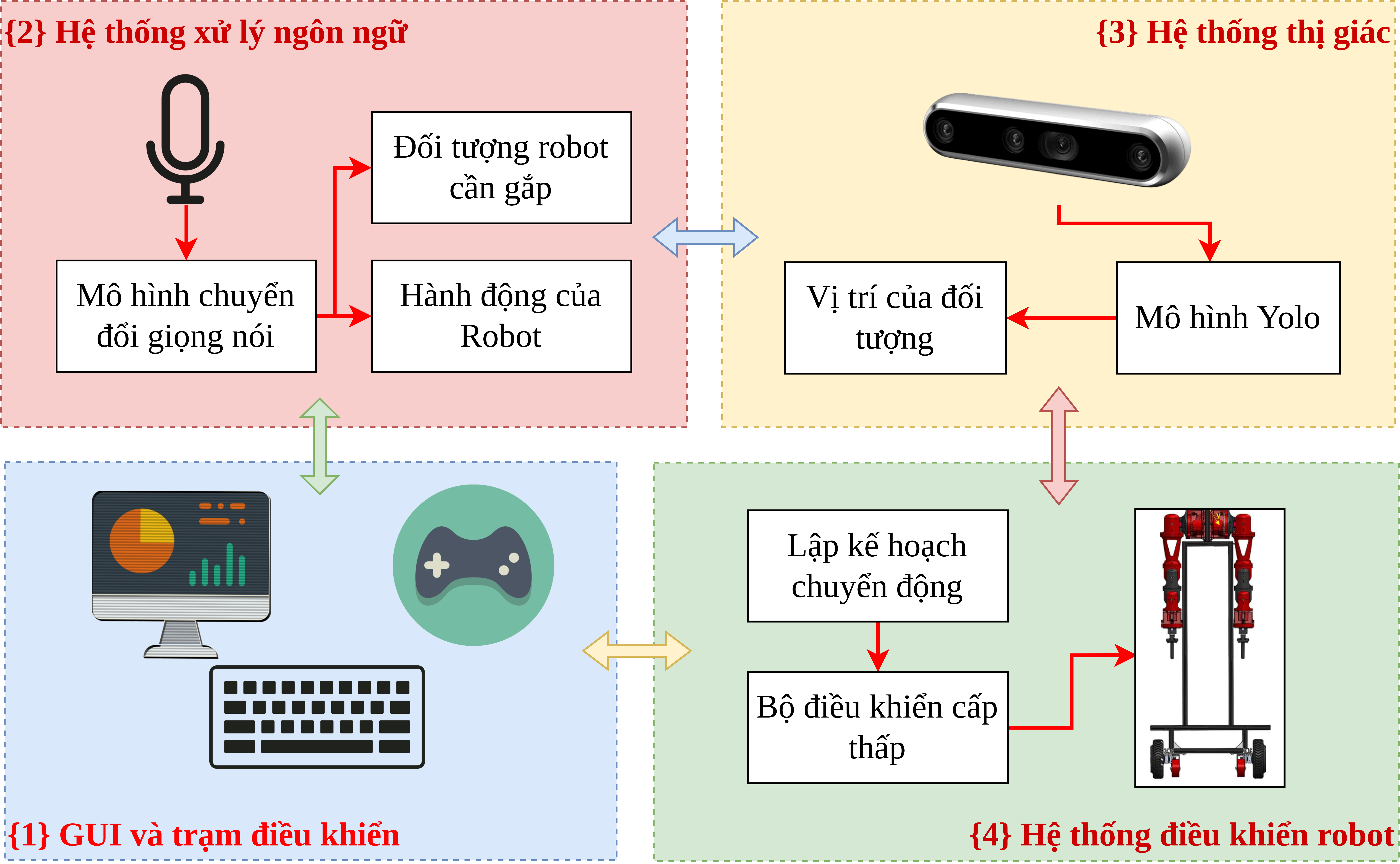}
    \caption{Sơ đồ tổng quan hệ thống tay máy đôi}
    \label{fig:overview}
\end{figure}

Hình \ref{fig:overview} mô tả tổng quan hệ thống tay máy đôi bao gồm bốn thành phần chính: \{1\} khối giao diện đồ họa người dùng (GUI) và điều khiển; \{2\} khối xử lý ngôn ngữ tự nhiên, chuyển thành các câu lệnh điều khiển tay máy đôi; \{3\} khối thị giác, xác định vật thể, tọa độ gắp vật và \{4\} khối điều khiển chuyển động của robot. Hệ động được thiết kế tối ưu và mô-dun hóa trong đó khối \{1\}, \{4\} được mô tả chi tiết ở phần \ref{sec:softwareROS} và khối \{2\}, \{3\} trình bày trong phần \ref{sec:multiAI}.

\subsection{Thiết kế nền tảng tay máy đôi}
\begin{figure}[!ht]
    \centering
    \includegraphics[width=0.38\textwidth]{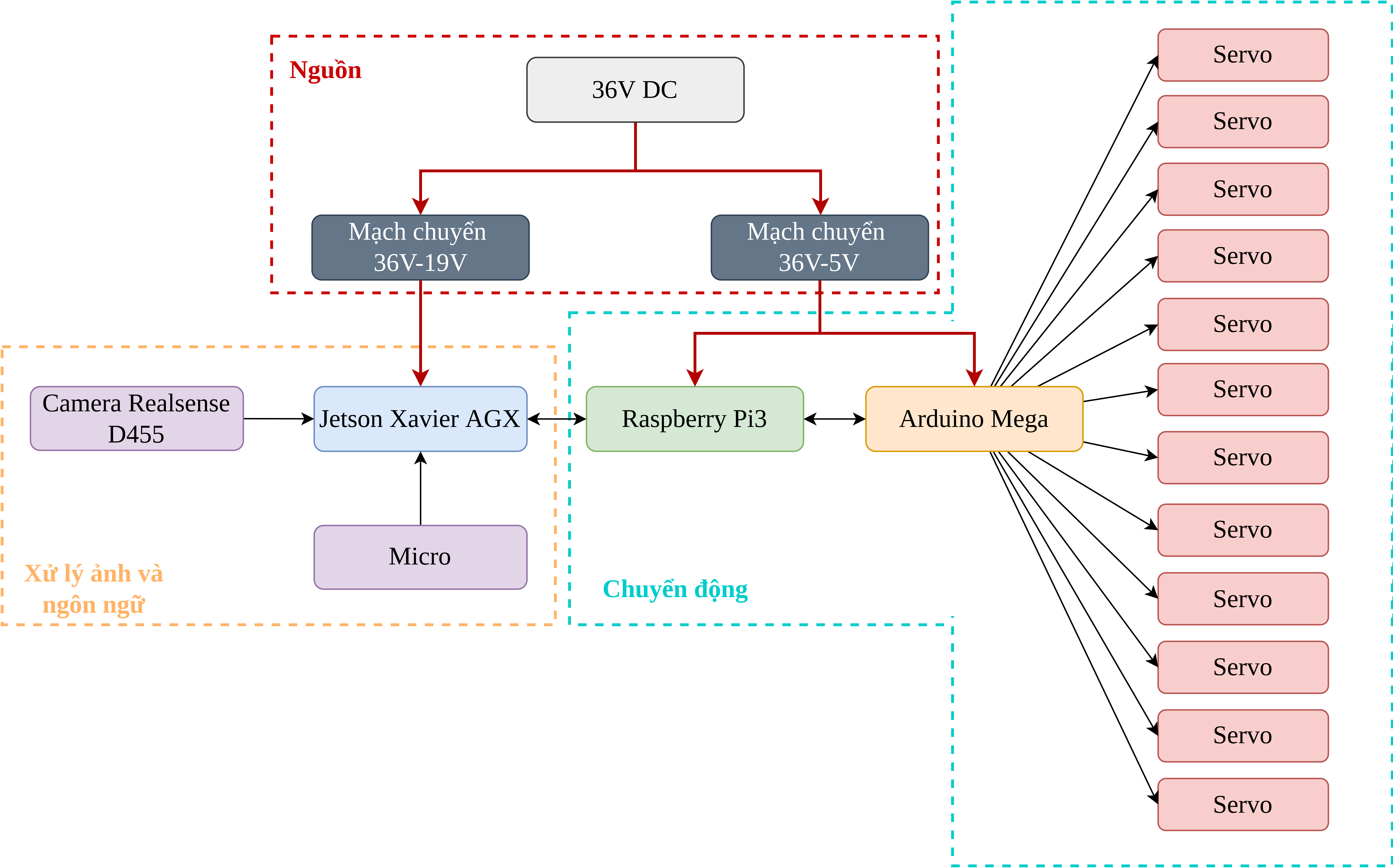}
    \caption{Sơ đồ kết nối các linh kiện của robot}
    \label{fig:component_diagram}
\end{figure}

Cấu trúc thiết kế của nền tảng robot tay máy đôi được thể hiện trong Hình \ref{fig:components} với vị trí chi tiết của các thiết bị. Hình \ref{fig:component_diagram} mô tả sơ đồ kết nối thành phần hệ thống trong đó robot được cấu thành bởi ba thành phần chính: 
\begin{figure}[!ht]
    \centering
    \includegraphics[width=0.38\textwidth]{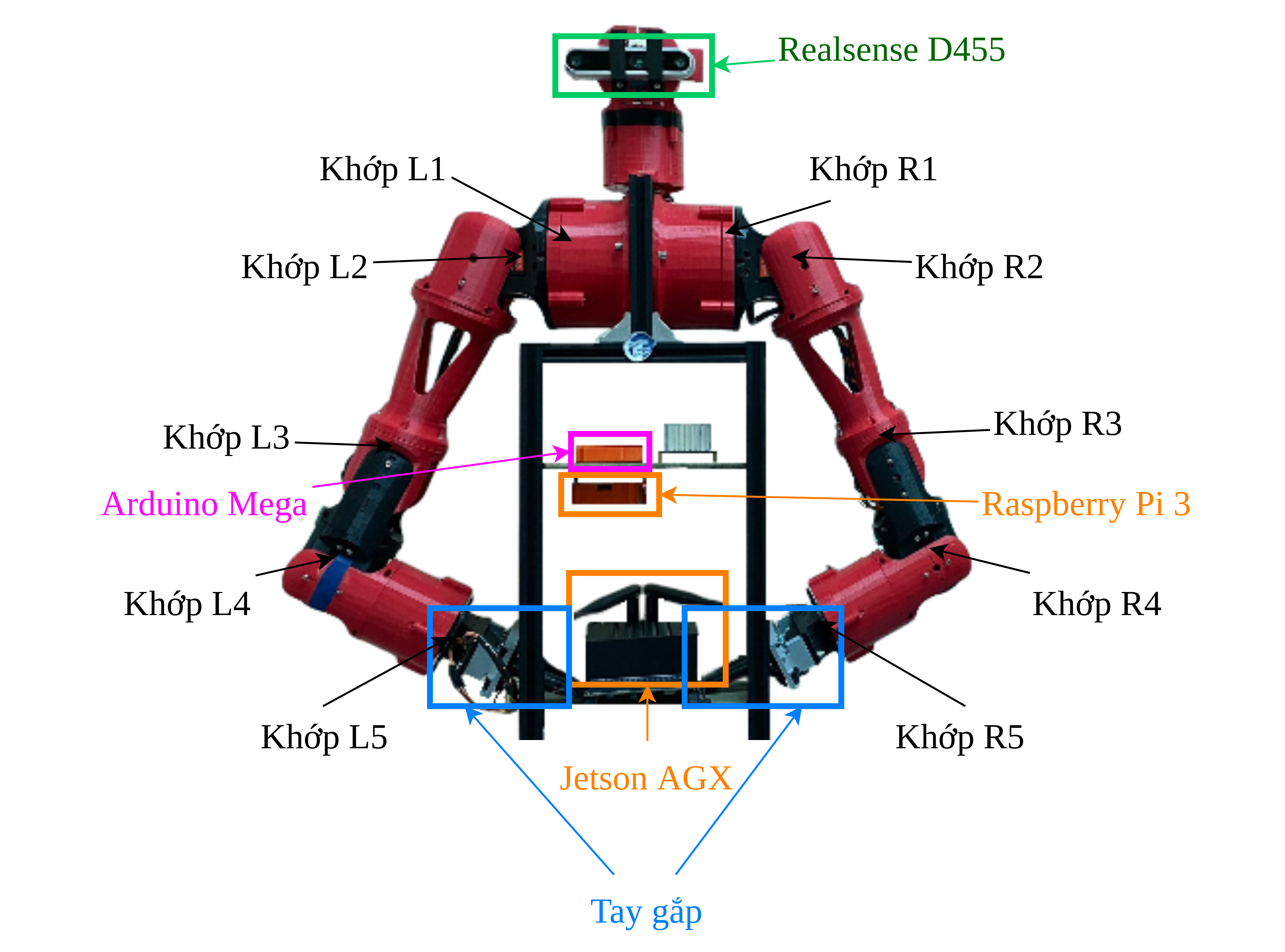}
    \caption{Vị trí các thiết bị trên khung Robot}
    \label{fig:components}
\end{figure}
\begin{itemize}
    \item Khối xử lý ảnh và xử lý ngôn ngữ - nhận thông tin từ môi trường thông qua camera realsene D455 và micro sau đó được xử lý thông tin qua máy tính nhúng Jetson Xavier AGX;
    \item  Khối chuyển động gồm 12 servo tương ứng với từng khớp và tay gắp, Arduino Mega chịu trách nhiệm điều khiển trực tiếp động cơ. Các thông tin được quản lý bởi Raspberry Pi thông qua các tính toán động học, động học ngược.
    \item Khối nguồn - cung cấp điện áp cho hai khối trên bao gồm một bộ chuyển đổi từ 36V xuống 5V cho khối chuyển động và bộ chuyển đổi từ 36V xuống 19V cho khối xử lý ảnh và ngôn ngữ.
\end{itemize}

Để mở rộng phạm vi chuyển động của robot, chúng tôi đã sử dụng các thanh nhôm định hình để gắn cánh tay cao hơn. Robot có kích thước 120cm x 40cm x 22cm tương ứng với chiều cao, chiều dài và chiều rộng trong đó mỗi cánh tay dài khoảng 78cm tính từ gốc tay tới đầu tay gắp. Robot được gia công bằng phương pháp in 3D với vật liệu nhựa với từng mô-đun riêng biệt nên dễ dàng lắp đặt, điều chỉnh đồng thời đáp ứng được các nhu cầu hoạt động nhẹ.

\begin{figure}[!ht]
    \centering
    \includegraphics[width=0.4\textwidth]{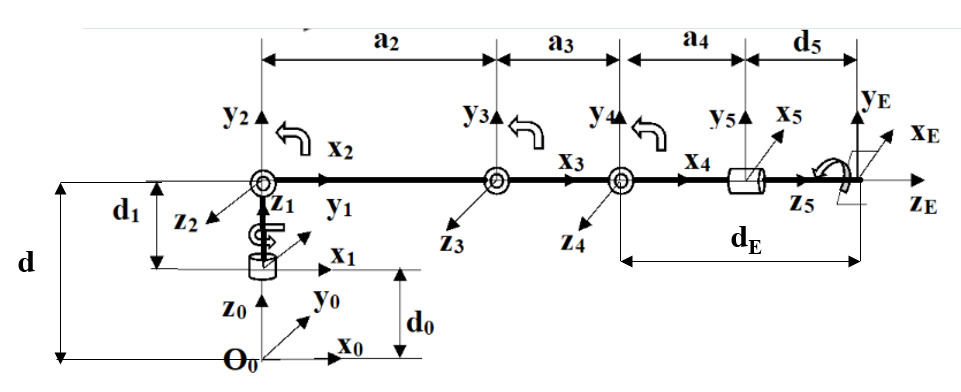}
    \caption{Hệ trục tọa độ của robot}
    \label{fig:coordinate}
\end{figure}

Hình \ref{fig:coordinate} là các trục tọa độ được gắn tại tâm của các khớp quay, trục $z$ là trục khâu tiếp theo quay quanh nó, trục $x$ thường được đặt dọc theo pháp tuyến chung và hướng từ khớp thứ $i$ đến $i+1$, trục y  được xác định theo quy tắc bàn tay phải.

Tiếp theo chúng tôi lập bảng tham số D-H và tính toán ma trận chuyển đổi cho từng khớp. Kết quả cuối cùng thu được phương trình động học theo phương trình (\ref{eq:kinematicequation}):

\begin{equation}
\label{eq:kinematicequation}
        T = A^0_1*A^1_2*A^2_3*A^3_4*A^4_5 \\
        =
        \begin{bmatrix}
            i_x & j_x & k_x & p_x \\
            i_y & j_y & k_y & p_y \\
            i_z & j_z & k_z & p_z \\
            0 & 0 & 0 & 1
        \end{bmatrix}
\end{equation}

trong đó, 

\hspace{1cm} $i_x = sin_1sin_5 + cos_5cos_1cos_{234}$

\hspace{1cm} $j_x = -sin_5cos_1cos_{234}-cos_5sin_1$

\hspace{1cm} $i_y=cos_1sin_5+cos_5sin_1cos_{234}$ 

\hspace{1cm} $J_y = cos_5sin_1-sin_5sin_1cos_{234}$

\hspace{1cm} $i_z = cos_5sin_{234}$ 

\hspace{1cm} $j_z = -sin_5sin_{234}$

\hspace{1cm} $k_x = -cos_1sin_{234}$

\hspace{1cm} $p_x=cos_1(a_2cos_2-d_Esin_{234}+a_3cos_{23})$

\hspace{1cm} $k_y = -sin_1sin_{234}$

\hspace{1cm} $p_y=sin_1(a_2cos_2 - d_Esin_{234}+a_3cos_{23})$

\hspace{1cm} $k_z = cos_{234}$ 

\hspace{1cm} $p_z=d+ a_2sin_2+a_3sin_{23}+d_Ecos_{234}$

Mục đích của bài toán động học nghịch đảo là tính toán các góc khớp của robot khi vị trí của hiệu ứng cuối đã được biết. Vector góc khớp được xác định theo công thức (\ref{eq:inverK}):

\begin{equation}
\label{eq:inverK}
    \begin{aligned}
        q = 
        \begin{bmatrix}
            \theta_1 & \theta_2 & \theta_3 & \theta_4 &\theta_5
        \end{bmatrix} ^T  \hspace{0.5cm}
    \end{aligned}
\end{equation}

Phương trình động học ngược \cite{Iliukhin2017} của robot thu được như sau: 
\begin{equation}
    \begin{aligned}
        \theta_1 = \tan^{-1}\frac{p_y}{p_x}
    \end{aligned}
\end{equation}

\begin{equation}
\label{eq:theta2}
    \begin{aligned}
        \theta_2 = \tan^{-1}{\frac{n(a_2+a_3cos_3)-ma_3sin_3}{ba_3sin_3+m(a_2+a_3cos_3)}}
    \end{aligned}
\end{equation}

\begin{equation}
\label{eq:theta3}
    \begin{aligned}
        \theta_3 = \cos^{-1}{\frac{m^2+ n^2-a^2_2-a^2_3}{2a_2a_3}}
    \end{aligned}
\end{equation}

\begin{equation}
    \begin{aligned}
        \theta_{234} =\tan^{-1}{\frac{a_2cos_2+a_3cos_{23}-p_xcos_1-p_ysin_1}{p_z - d-a_2sin_2-a_3sin_{23}}}
    \end{aligned}
\end{equation}

\begin{equation}
    \begin{aligned}
        \theta_3 = \theta_{234} - \theta_2 - \theta_3
    \end{aligned}
\end{equation}

\begin{equation}
    \begin{aligned}
        \theta_5 = \tan^{-1}{\frac{(i_ycos_1-i_xsin_1)sin_{234}}{i_ysin_1+i_xcos_1}}
    \end{aligned}
\end{equation}

Trong phương trình \ref{eq:theta2} và \ref{eq:theta3} $m$, $n$ được định nghĩa theo phương trình (\ref{eq:m}), (\ref{eq:n}):

\begin{equation}
\label{eq:m}
    \begin{aligned}
        m = p_xcos_1 + p_ysin_1+d_Esin_{234}
    \end{aligned}
\end{equation}

\begin{equation}
\label{eq:n}
    \begin{aligned}
        n = p_z d - d_Ecos_{234}
    \end{aligned}
\end{equation}

\subsection{Thiết kế khung điều khiển phần mềm dựa trên ROS}
\label{sec:softwareROS}
\begin{figure}[!hb]
    \centering
    \includegraphics[width=0.4\textwidth]{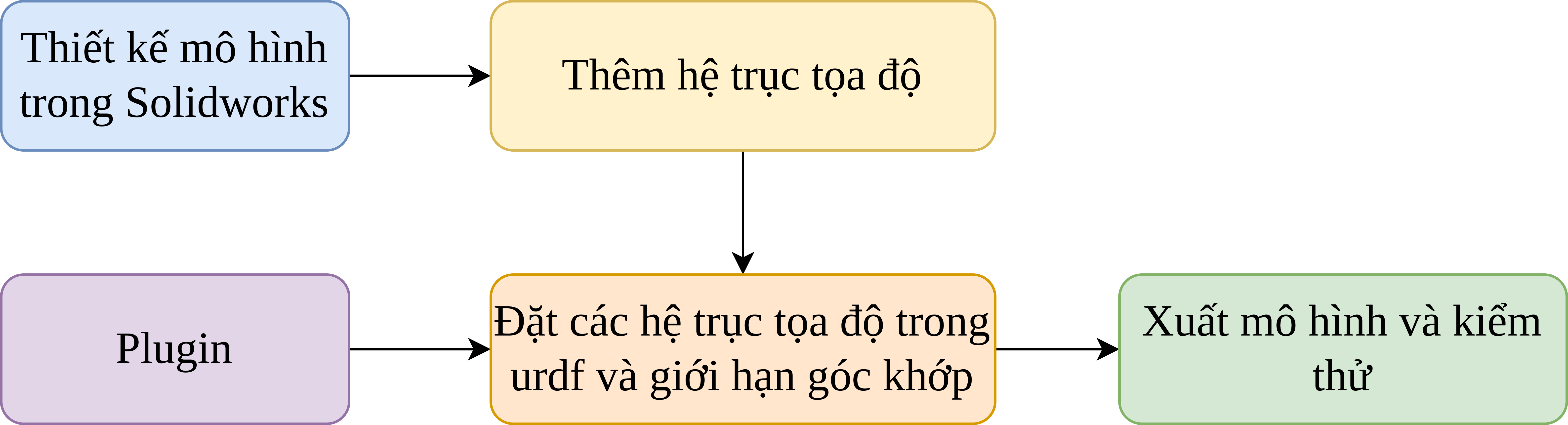}
    \caption{Các bước tạo mô hình URDF từ mô hình Solidwork}
    \label{fig:urdf}
\end{figure}

\begin{figure}[!hb]
    \centering
    \includegraphics[width=0.4\textwidth]{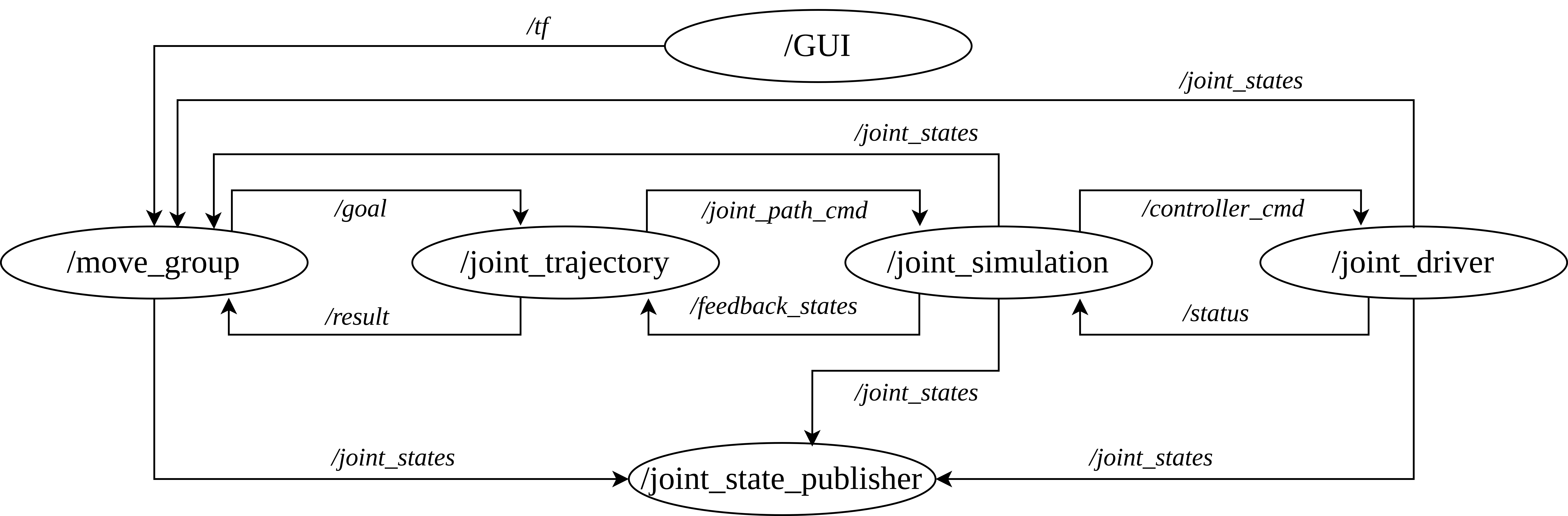}
    \caption{Mô hình mối quan hệ giữa các node trong ROS}
    \label{fig:moveit}
\end{figure}

\begin{figure}[!hb]
    \centering
    \includegraphics[width=0.2\textwidth]{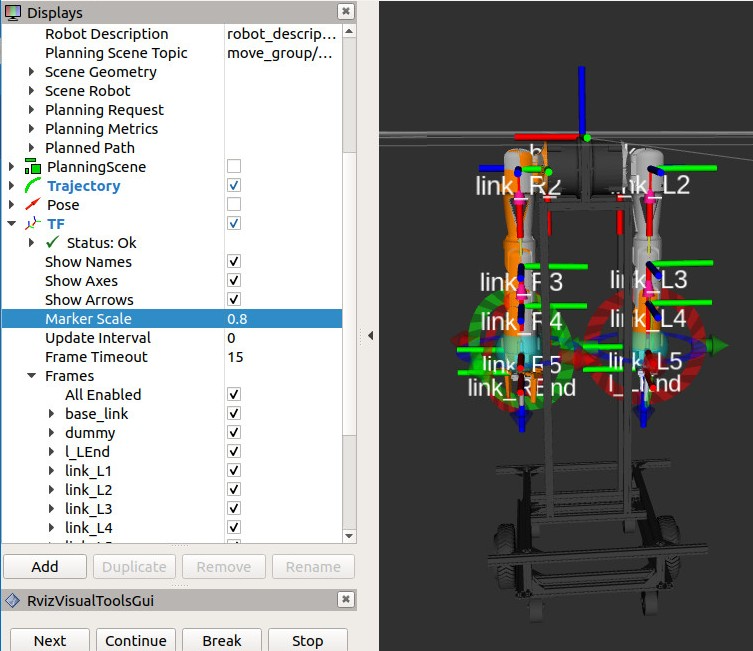}
    \caption{Trực quan hóa mô hình robot trong ROS}
    \label{fig:Ros_model}
\end{figure}

Trong bài báo này, chúng tôi sử dụng Moveit là công cụ chính cho các tác vụ lập kế hoạch chuyển động, động học, động học ngược của robot. Về cơ bản, khung điều khiển bao gồm ba bước chính: Thiết lập mô hình URDF (Unified Robot Description Format) nhằm trực quan hóa robot trong môi trường mô phỏng, các bước được mô tả chi tiết qua Hình \ref{fig:urdf}. Tiếp theo, chúng tôi triển khai bộ điều khiển cấp thấp của robot trong ROS thông qua Arduino, cuối cùng chúng tôi tạo các bộ cấu hình trong Moveit trình bày trong Hình \ref{fig:moveit}. Hình \ref{fig:Ros_model} trực quan hóa tập mô hình URDF thông qua RViz.

Chúng tôi tạo node \textit{/GUI} nhằm tiếp nhận câu lệnh, hiển thị thông tin robot. Sau đó câu lệnh tới node \textit{\/move\_group} thực hiện các tác vụ như lập kế hoạch chuyển động, thiết lập các khớp. Node \textit{/joint\_trajectory} hiển thị giao diện hành động với độ điều khiển quỹ đạo. Sau đó, thông tin về quỹ đạo được cấp tới node \textit{/joint\_simulator} điều khiển mô phỏng robot tay máy đôi. Cuối cùng, giá trị động cơ (góc khớp) được truyền xuống bộ điều khiển cấp thấp qua cầu nối node \textit{joint\_driver} giữa \textit{ros\_controller} và \textit{robot\_controller}.

\subsection{Trí tuệ nhân tạo đa thể thức}
\label{sec:multiAI}
\subsubsection{Nhận dạng đối tượng}

\begin{figure}[!ht]
    \centering
    \includegraphics[width=0.4\textwidth]{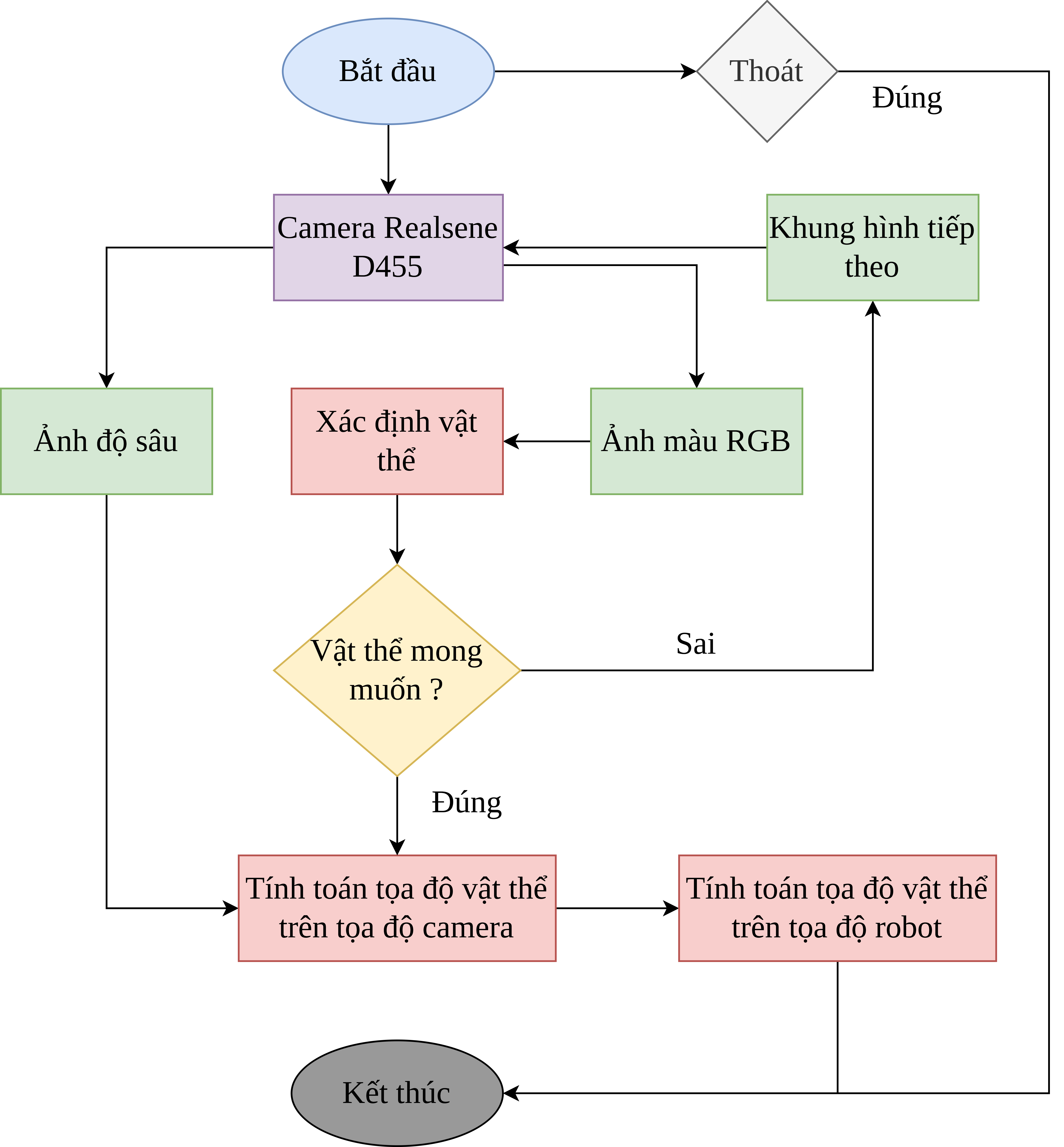}
    \caption{Nhận dạng đối tượng và ước tính tọa độ vật thể}
    \label{fig:coordinate_estimate}
\end{figure}

Hình \ref{fig:coordinate_estimate} trình bày chi tiết các bước nhận dạng đối tượng đồng thời xác định tọa độ trong không gian làm việc của robot. Thông tin từ môi trường được trích xuất thông qua máy ảnh chiều sâu realsense 455. Sau đó, dữ liệu ảnh mày RGB được sử dụng làm đầu vào của mô hình nhận dạng vật thể. Trong bài báo này, chúng tôi sử dụng mô hình Yolo (You only look once) \cite{Redmon2015} cho bài toán xác định đối tượng. Đầu ra của mô hình cung cấp chiều dài, chiều rộng và tọa độ tâm của đối tượng trong miền pixel 2D của ảnh. 

Sau khi xác nhận được vật thể mong muốn, chúng tôi bổ sung thông tin độ sâu nhằm ước tính tọa độ vật thể trong môi trường thực. Đầu tiên chúng tôi tính toán tọa độ vật thể trên khung tọa độ camera theo hai bước \cite{Andriyanov2022}:
\begin{itemize}
    \item B1: Xác định tọa độ của một điểm ảnh từ một cảnh: chúng tôi sử dụng phép biến đổi hình chiếu phối cảnh được mô tả theo Hình \ref{fig:prjTransf}. Vấn đề đặt ra là cần xác định tọa độ điểm chiếu trên mặt phẳng ảnh. Xét điểm ảnh $(x_i, y_i)$ có tọa độ thực thế $(X_s, Y_s, Z_s)$. Theo quy tắc đồng dạng của tam giác ta có phương trình chiếu như sau:
\begin{equation}
    \begin{aligned}
    \biggl\{ 
        \begin{array}{*{20}{c}}
            x_i = f \frac{X_s}{Z_s}\\
            y_i = f \frac{Y_s}{Z_s}
        \end{array}
    \end{aligned}   
\end{equation}

\begin{figure}[!hb]
    \centering
    \includegraphics[width=0.35\textwidth]{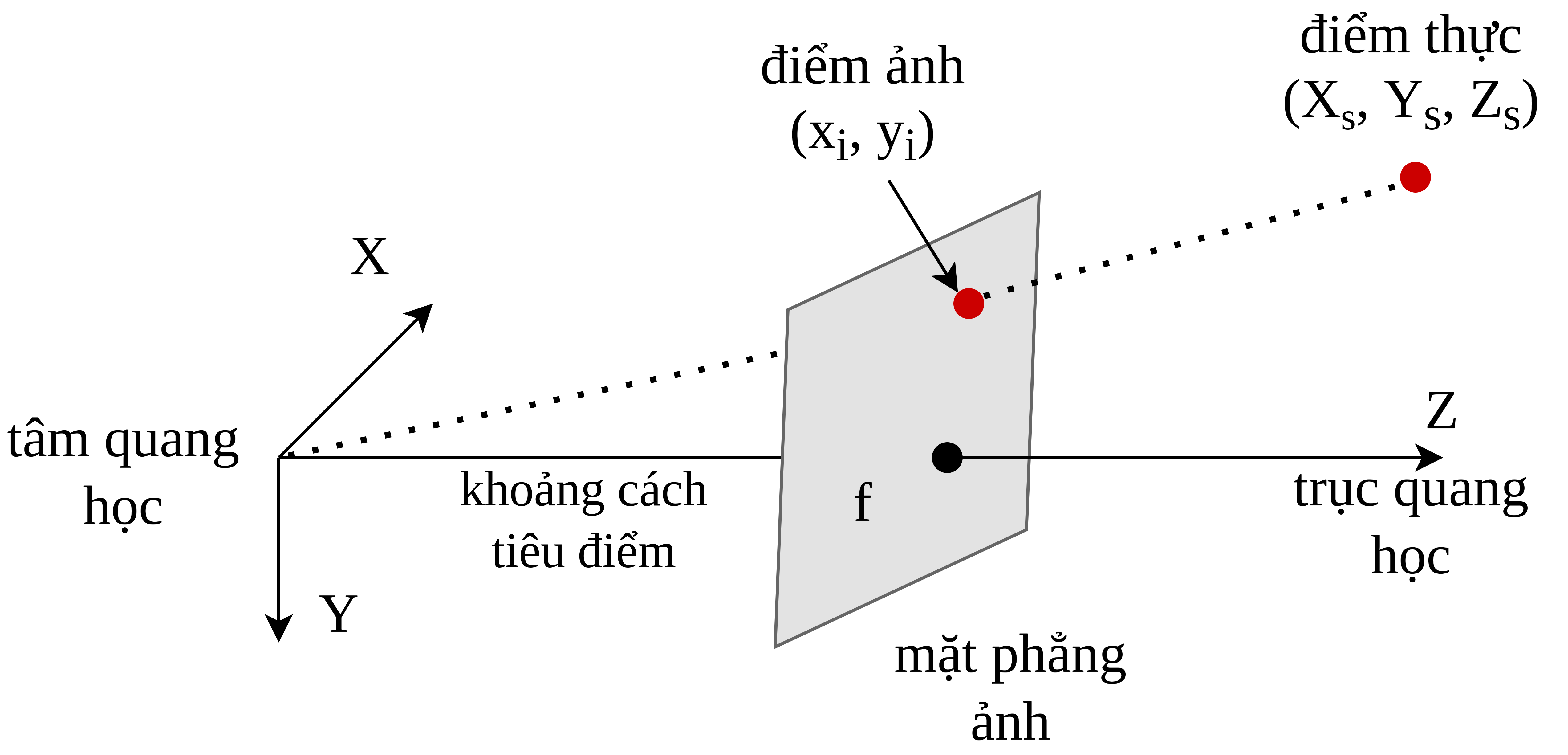}
    \caption{Mô hình phép biến đổi hình chiếu phối cảnh}
    \label{fig:prjTransf}
\end{figure}

\item B2: Chuyển đổi qua hệ tọa độ hình ảnh với tâm quang học $(c_x, c_y)$ như Hình \ref{fig:imgCoordinate}

\begin{equation}
    \begin{aligned}
    \biggl\{ 
        \begin{array}{*{20}{c}}
            u = f \frac{X_s}{Z_s} + c_x\\
            v = f \frac{Y_s}{Z_s} + c_y
        \end{array}
    \end{aligned}   
\end{equation}

Hệ tọa độ thực $(X_s, Y_s)$ của vật thể thu được:
\begin{equation}
    \begin{aligned}
    \biggl\{ 
        \begin{array}{*{20}{c}}
            X_s = Z_s \frac{u- c_x}{f} \\
            Y_s = Z_s \frac{v-c_y}{f}
        \end{array}
    \end{aligned}   
\end{equation}
với $Z_s$ có thể thu được khi sử dụng máy ảnh độ sâu tương ứng với tọa độ tâm thu được.

\begin{figure}[!hb]
    \centering
    \includegraphics[width=0.25\textwidth]{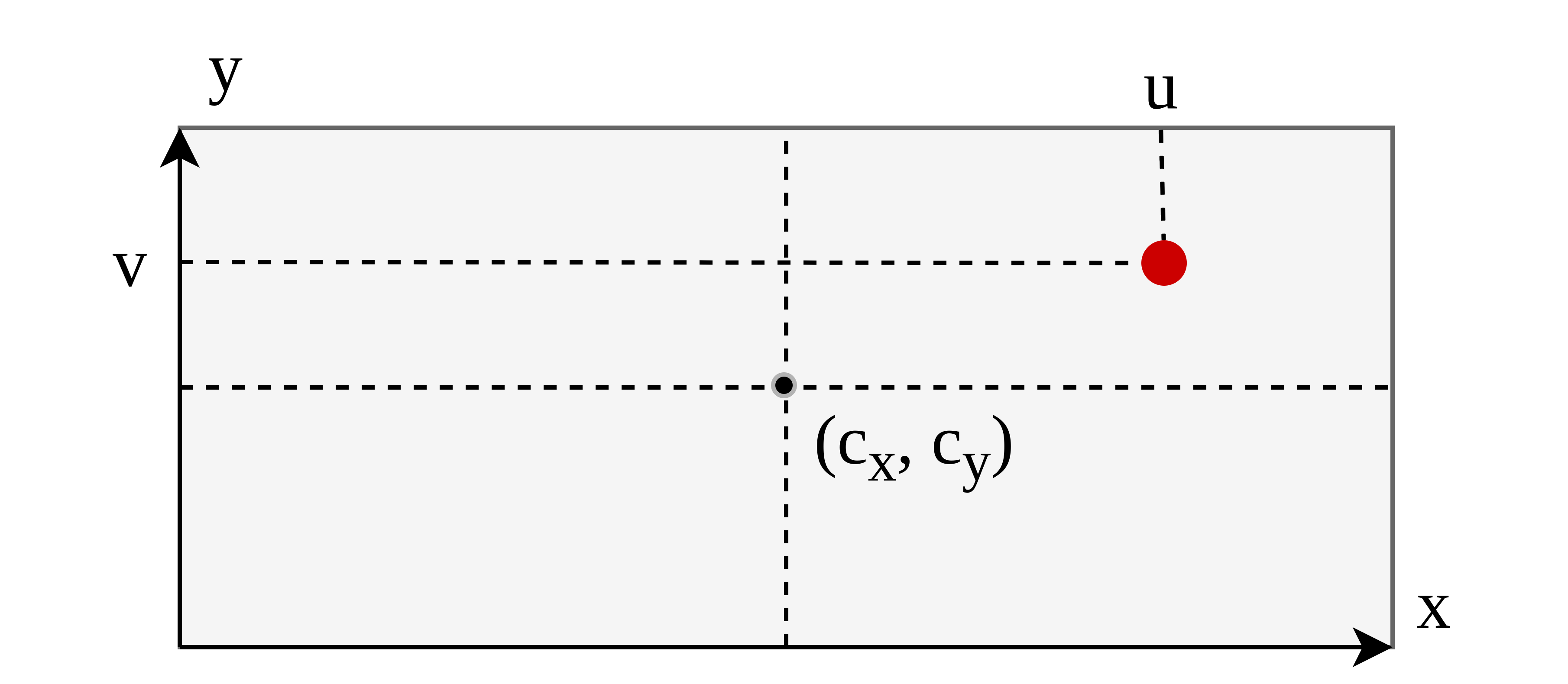}
    \caption{Biểu diễn hệ toạ độ hình ảnh}
    \label{fig:imgCoordinate}
\end{figure}
\end{itemize}


Tiếp theo, chúng tôi tính toán tọa độ vật thể trên hệ toạ độ của robot. Ta có hệ tọa độ $(X_c, Y_c, Z_c)$ của camera, hệ tọa độ $(X_r, Y_r, Z_r)$ của robot và ma trận ${ }^{r} \mathbf{P}_{c}$ là ma trận chuyển đổi giữa hệ tọa độ camera qua hệ toạ độ robot được xác định bởi:
\begin{equation}
\begin{aligned}
\begin{bmatrix}
X_{r} \\
Y_{r} \\
Z_{r} \\
1
\end{bmatrix} &={ }^{r} \mathbf{P}_{c}
\begin{bmatrix}
X_{c} \\
Y_{c} \\
Z_{c} \\
1
\end{bmatrix} 
=\begin{bmatrix}
{ }^{r} \mathbf{R}_{c} & { }^{r} \mathbf{T}_{c} \\
0_{1 \times 3} & 1
\end{bmatrix}
\begin{bmatrix}
X_{c} \\
Y_{c} \\
Z_{c} \\

\end{bmatrix},
\end{aligned}
\end{equation}

trong đó ${ }^{r} \mathbf{R}_{c}$ và ${ }^{r} \mathbf{T}_{c}$ lần lượt là ma trận xoay và ma trận dịch chuyển giữa hệ tọa độ camera và hệ tọa độ robot

\subsubsection{Nhận dạng giọng nói}
\begin{figure}[!hb]
    \centering
    \includegraphics[width=0.35\textwidth]{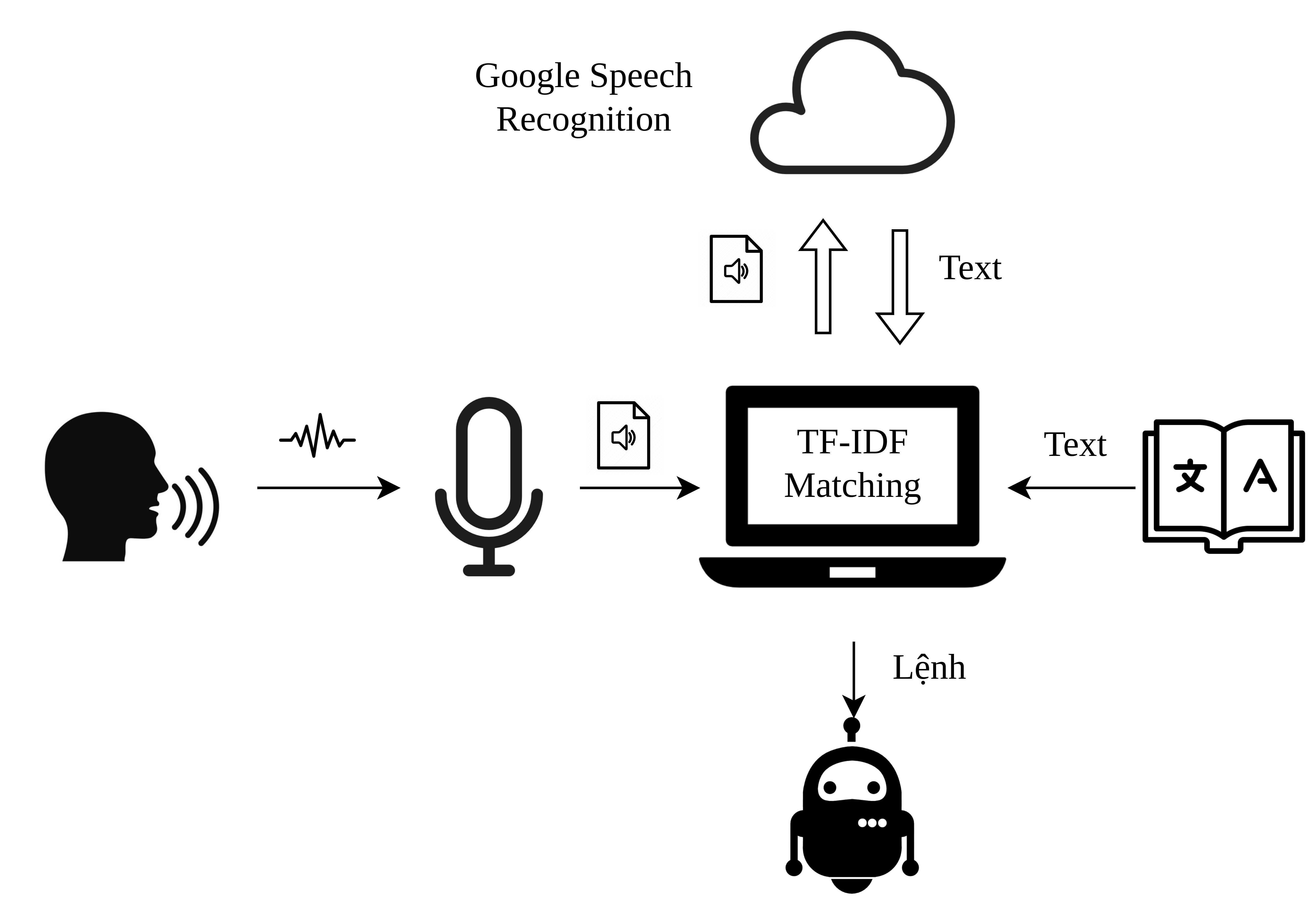}
    \caption{Mô hình nhận dạng giọng nói}
    \label{fig:speechRecog}
\end{figure}

Mô hình nhận dạng giọng nói được thể hiện thông qua Hình \ref{fig:speechRecog}. Giọng nói được thu thập thông qua micro sau đó lưu trữ dạng các tập tin âm thanh. Tiếp theo, tập tin âm thanh được đi qua khối Google Speech Recognition nhằm chuyển đổi giọng nói thành văn bản. Chúng tôi thiết lập bộ từ điển nhằm đưa ra các giả thiết khả dĩ về hành vi, chức năng của robot. Sau khi có dữ liệu văn bản của khối hận dạng giọng nói và khối từ điển, chúng tôi thực hiện so khớp hai thông tin nhờ TF-IDF \cite{qaiser2018text, minh2018} và phương trình Cosine Simalarity \cite{Gunawan2018}. TF-IDF tiếp nhận văn bản sau đó dựa vào tần số xuất hiện của từ trong câu thể hiện tầm quan trong của một từ ngữ từ đó tạo ra vector đặc trưng cho mỗi câu theo công thức:

\begin{equation}
    \begin{aligned}
        tf idf(t, d, D) = tf(t, d) . idf(t, D)
    \end{aligned}
\end{equation}
trong đó, $tf(t, d)$ thể hiện tần số xuất hiện của từ $w$ trong câu $d$. $idf(t, D) = \log\frac{N}{count(d \in D: t \in d)}$ thể hiện mức độ phổ biến của từ  với $N$ số lượng câu $d$ trong tập thư viện $D$.
    
Cuối cùng, câu lệnh được trích xuất và truyền tới robot.

\section{THỰC NGHIỆM VÀ ĐÁNH GIÁ KẾT QUẢ}
\label{Sec:KetQuaMoPhong}

\begin{table}[ht]
\caption{Tập dữ liệu mô hình nhận dạng vật thể, giọng nói}
\centering
\begin{tabular}{m{0.25\textwidth} |m{0.06\textwidth} |m{0.06\textwidth}}
\hline
Câu lệnh điều khiển & \begin{center}
    Đối tượng
\end{center} & \begin{center} Nhãn Yolo \end{center}\\
\hline
pick up the white rectangular object & \begin{center}\ \includegraphics[width=0.05\textwidth]{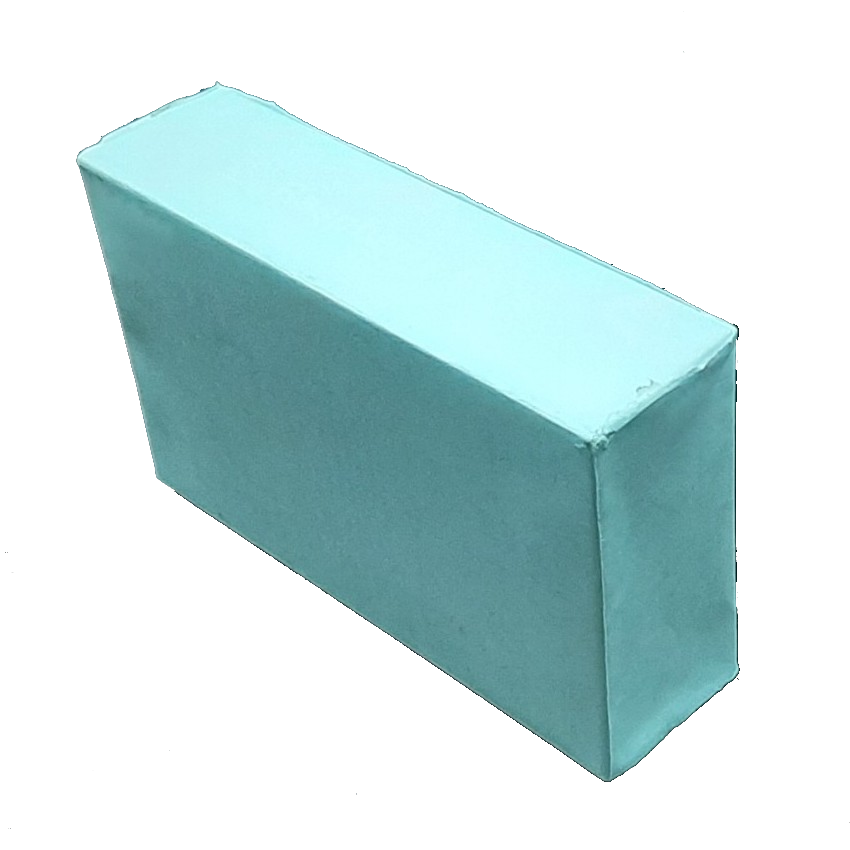}\end{center} & \begin{center}
    rectangle
\end{center}\\ 
\hline

pick up the white cylinder object & \begin{center}\ \includegraphics[width=0.05\textwidth]{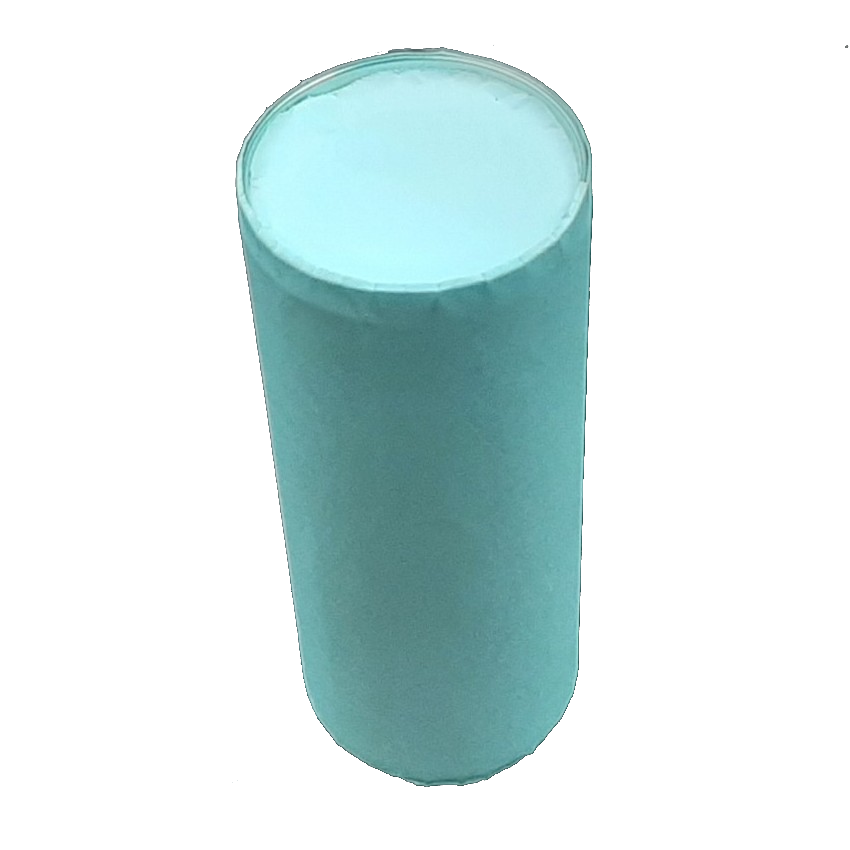}\end{center} & \begin{center}
    cylinder
\end{center}\\ 
\hline

pick up the box & \begin{center}\ \includegraphics[width=0.05\textwidth]{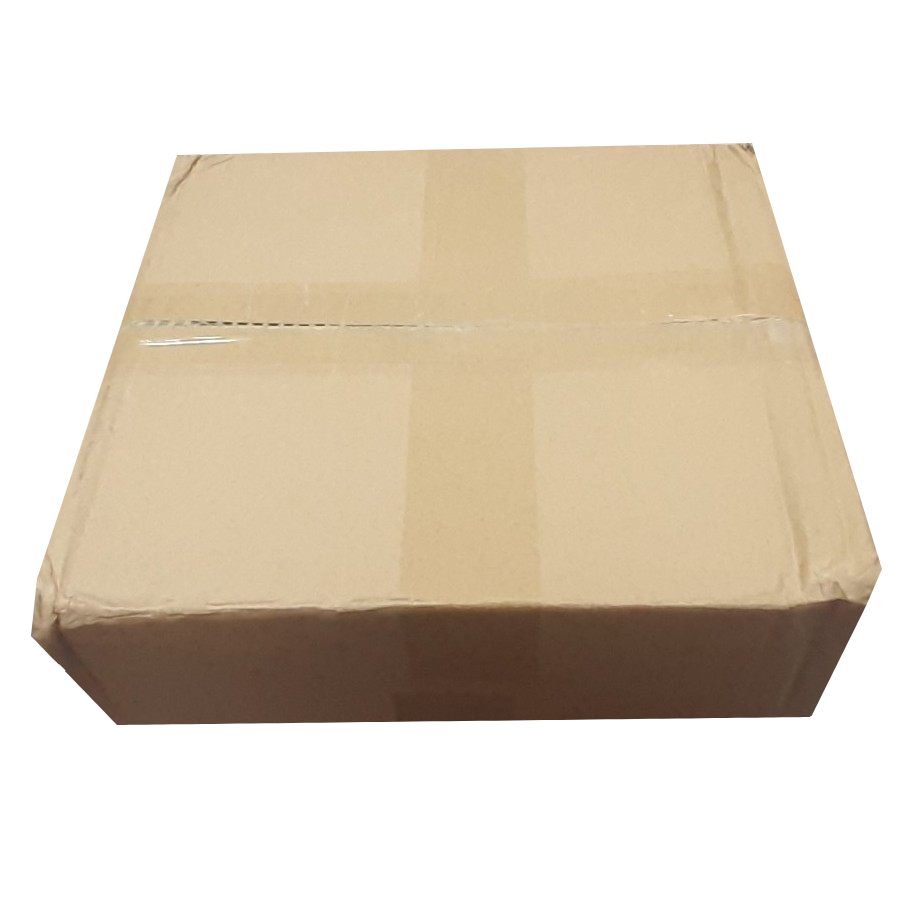}\end{center} & \begin{center}
    box
\end{center}\\ 
\hline
\end{tabular}
\label{Tab:dataset}
\end{table}

\begin{table*}[ht]
\caption{Kết quả sai số hệ thống (đơn vị: cm)}
\centering
\begin{tabular}{m{0.12\textwidth} m{0.04\textwidth} m{0.04\textwidth} m{0.04\textwidth} m{0.04\textwidth} m{0.04\textwidth} m{0.04\textwidth} m{0.04\textwidth} m{0.04\textwidth} m{0.04\textwidth} m{0.04\textwidth} m{0.06\textwidth}}
\hline
Lần đo & 1 & 2 & 3 & 4 & 5 & 6 & 7 & 8 & 9 & 10 & TB \\ \hline
Tay trái & 1.05 & 1.28 & 1.29 & 1.55 & 1.56 & 1.25 & 0.88 & 1.86 & 0.78 & 1.48 & \textbf{1.30} \\ \hline
Tay phải & 1.07 & 0.88 & 0.96 & 1.36 & 1.68 & 1.24 & 1.87 & 0.88 & 1.5 & 1.4 & \textbf{1.28} \\ \hline
Kết hợp Yolo &  1.3 & 1.24 & 1.84 & 3.48 & 3.82 & 1.51 & 0.41 & 0.97 & 1.5 & 2.26 & \textbf{1.83} \\
\hline
\end{tabular}
\label{Tab:resultsystem}
\end{table*}

\subsection{Điều kiện đánh giá}

\begin{figure}[!hb]
    \centering
    \includegraphics[width=0.3\textwidth]{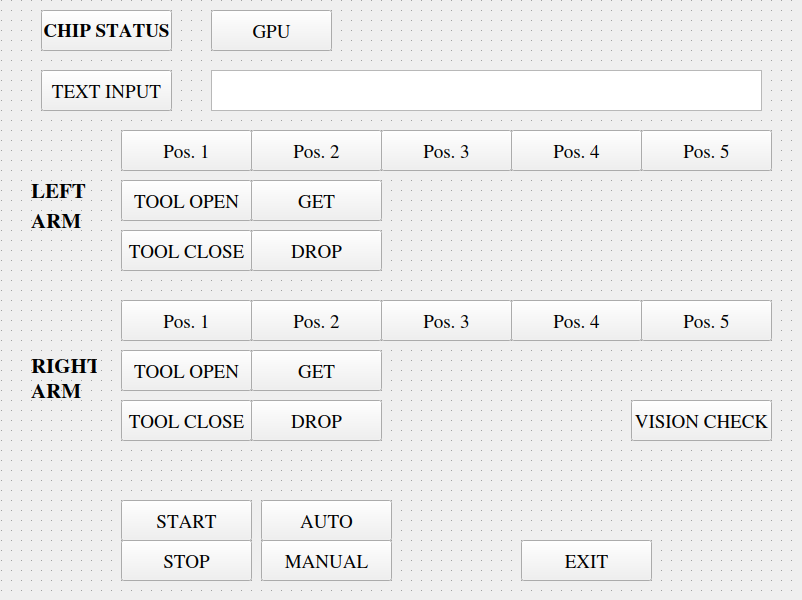}
    \caption{Giao diện điều khiển hệ thống}
    \label{fig:gui}
\end{figure}
Hệ thống được thử nghiệm và đánh giá trong môi trường thực trong đó bộ tham số góc khớp bị giới hạn với gia tốc tối đa 0.1 ($rad/s^2$) và vận tốc tối đa trong khoảng 1.73 đến 2.56 ($rad/s$). Bộ dữ liệu cho quá trình nhận dạng vật thể bao gồm 100 ảnh kích cỡ 640x480 chứa các đối tượng được mô tả như trong Bảng \ref{Tab:dataset} cùng với câu lệnh điều kiểu của chúng. Giao diện điều khiern hệ thống được chúng tôi thiết kế như Hình \ref{fig:gui} bao gồm các câu lệnh điều khiển cơ bản, hiển thị câu lệnh. Sai số hệ thống được đánh giá thông qua phương pháp so sánh tọa độ thực và tọa độ tính toán, sai số là khoảng cách giữa hai điểm theo phương trình Euclid:
\begin{equation}
    \begin{aligned}
        err = \sqrt{(x - x_r)^2 + (y-y_r)^2+(z-z_r)^2}
    \end{aligned}
\end{equation}

trong đó: $(x, y, z)$ là tọa độ mong muốn, $(x_r, y_r, z_r)$ là tọa độ thực tế.

\subsection{Kiểm tra độ chính xác của mô hình nhận dạng vật kể và giọng nói}
Chúng tôi tiến hành kiểm thử mô hình nhận dạng giọng nới với ba câu lệnh được mô tả như Bảng \ref{Tab:dataset}, trong đó mỗi câu lệnh được lặp lại 200 lần. Kết quả từ Bảng \ref{Tab:resultspeech} cho thấy mô hình nhận dạng giọng nói cho kết quả tương đối chính xác ($> 97 \%$). Tuy nhiên để có thể ứng dụng vào trong công nghiệp, độ chính xác đạt được cần phải ở mức cao hơn như $99\%$.

\begin{table}[ht]
\caption{Kết quả thử nghiệm nhận dạng giọng nói}
\centering
\begin{tabular}{m{0.13\textwidth} |m{0.13\textwidth}| m{0.12\textwidth}}
\hline
pick up the white rectangular object & pick up the white cylinder object & pick up the box\\
\hline
0.975& 0.990 & 0.990 \\
\hline
\end{tabular}
\label{Tab:resultspeech}
\end{table}

\begin{figure}[!ht]
    \centering
    \includegraphics[width=0.3\textwidth]{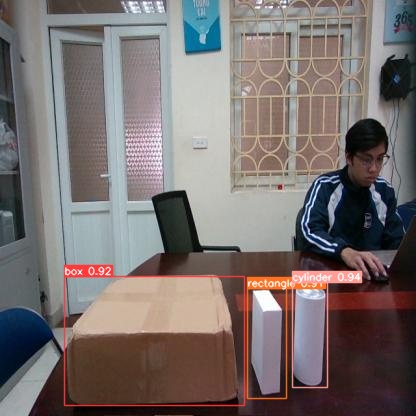}
    \caption{Kết quả nhận dạng vật thể}
    \label{fig:yolo}
\end{figure}

Mô hình nhận dạng vật thể bằng Yolo cho kết quả rất chính xác được thể hiện qua Hình \ref{fig:yolo}. Kết quả cho thấy các chỉ số như mAP\_0.5 đạt $95.71 \%$, mAP\_0.95 đạt $78.26 \%$, pecision đạt $95.67 \%$ và recall đạt $87.95 \%$. Bên cạnh đó, tập dữ liệu có quy mô nhỏ và các lỗi xảy ra trong quá trình gán nhãn ảnh hưởng trực tiếp tới kết quả nhận dạng.

\subsection{Kiểm tra độ chính xác của hệ thống}
Kết quả của Bảng \ref{Tab:resultsystem} cho thấy về mặt lý tưởng robot có thể di chuyển tới các ví trí trong không gian với dung sai xác định trước tuy nhiên trong thực tế có thêm sai số động cơ và sai số của phép đo nên tổng sai số của từng tay dao động trong khoảng $0.78 cm$ đến $1.86 cm$ đối với tay phải và $0.87 cm$ đến $1.87 cm$ với tay trái. Đối sai số khí gắp vât thông qua mô hình yolo, chúng tôi nhận thấy rằng sai số tọa độ tỉ lệ thuận với khoảng cách trục $x$ và trục $z$ với sai số nằm trung bình khoảng $1.83 cm$.

\section{KẾT LUẬN}
\label{Sec:KetLuan}
Bài báo trình bày hệ thống phát triển tay máy đôi cho tác vụ tương tác người-robot dựa trên hệ điều hành ROS. Các mô hình trí tuệ nhân tạo được cung cấp nhằm giải quyết các vấn đề cơ bản từ đó tạo một bộ khung hoàn chỉnh. Kết quả thu được chứng minh tính khả thi và khả năng ứng dụng vào thực tiễn đặc biệt là trong thời điểm dịch bệnh đang diễn ra.

Tuy nhiên, hệ thống còn nhiều thiếu sót về cả phần cứng và các tác vụ tương tác. Trong tương lai, chúng tôi sẽ cải thiện hệ thống phần cứng đồng thời thêm các mô hình như nhận dạng cử chỉ, cảm xúc, ... cho robot.
%

\bibliographystyle{IEEEtran}
\balance
\bibliography{reference}
\end{document}